\documentclass[11pt,a4paper]{article}
\usepackage[hyperref]{acl2018}
\usepackage{times}
\usepackage{latexsym}

\usepackage{url}

\usepackage{algorithm}% http://ctan.org/pkg/algorithms
\usepackage{algpseudocode}% http://ctan.org/pkg/algorithmicx

\usepackage{graphicx}
\usepackage{amsmath}
\DeclareMathOperator*{\argmax}{argmax}

\algblock{ParFor}{EndParFor}
\algnewcommand\algorithmicparfor{\textbf{parfor}}
\algnewcommand\algorithmicpardo{\textbf{do}}
\algnewcommand\algorithmicendparfor{\textbf{end\ parfor}}
\algrenewtext{ParFor}[1]{\algorithmicparfor\ #1\ \algorithmicpardo}
\algrenewtext{EndParFor}{\algorithmicendparfor}

\aclfinalcopy % Uncomment this line for the final submission
\def\aclpaperid{***} %  Enter the acl Paper ID here

\title{Fast Neural Machine Translation Implementation}  

  \author{Hieu Hoang$^{\dagger}$ \, Tomasz Dwojak$^{*}$ \, Rihards Krislauks$^{\ddagger}$ \\ {\bf Daniel Torregrosa$^{\mathparagraph}$ \, Kenneth Heafield$^{\dagger}$ } \\[2mm]
  $^{\dagger}$University of Edinburgh \, $^{*}$Adam Mickiewicz University \\ $^{\ddagger}$Tilde \,  $^{\mathparagraph}$Universitat d'Alacant }
  
\date{}

\begin{document}
\maketitle
\begin{abstract}

This paper describes the submissions to the efficiency track for GPUs at the Workshop for Neural Machine Translation and Generation by members of the University of Edinburgh, Adam Mickiewicz University, Tilde and University of Alicante. We focus on efficient implementation of the recurrent deep-learning model as implemented in Amun, the fast inference engine for neural machine translation. We improve the performance with an efficient mini-batching algorithm, and by fusing the softmax operation with the k-best extraction algorithm. Submissions using Amun were first, second and third fastest in the GPU efficiency track.

\end{abstract}

\section{Introduction}
\label{sec:Introduction}

As neural machine translation (NMT) models have become the new state-of-the-art, the challenge is to make their deployment efficient and economical. This is the challenge that this shared task~\citep{birch2018wnmt} is shining a spotlight on.

One approach is to use an off-the-shelf deep-learning toolkit to complete the shared task where the novelty comes from selecting the appropriate models and tuning parameters within the toolkit for optimal performance. % such as Tensorflow\footnote{https://www.tensorflow.org/} or Pytorch\footnote{https://pytorch.org/} 

We take an opposing approach by eschewing model selection and parameter tuning in favour of efficient implementation. We use and enhanced a custom inference engine, Amun~\citep{junczys2016neural}, which we developed on the premise that fast deep-learning inference is an issue that deserves dedicated tools that are not compromised by competing objectives such as training or support for multiple models. As well as delivering on the practical goal of fast inference, it can serve as a test-bed for novel ideas on neural network inference, and it is useful as a means to explore the upper bound of the possible speed for a particular model and hardware. That is, Amun is an inference-only engine that supports a limited number of NMT models that put fast inference on modern GPU above all other considerations. %The consequence of this focused approach can be seen in the results of the shared task ???.

We submitted two systems to this year's shared task for the efficient translation on GPU. Our first submission was tailored to be as fast as possible while being above the baseline BLEU score. Our second submission trades some of the speed of the first submission to return better quality translations.

%We will describe below our experimental setup as well as the enhancements to Amun that we feel enabled it to achieve its speed.

\section{Improvements}

We describe the main enhancements to Amun since the original 2016 publication that has improved translation speed.

\subsection{Batching}

The use of mini-batching is critical for fast model inference. The size of the batch is determined by the number of inputs sentences to the encoder in an encoder-decoder model. However, the number of batches during decoding can vary as some sentences have completed translating or the beam search add more hypotheses to the batch.

It is tempting to ignore these considerations, for example, by always decoding with a constant batch and beam size and ignoring hypotheses which are not needed. Figure~\ref{algo:simple-mini-batching} illustrates a na\"ive mini-batching with a constant size batch. The downside to this algorithm is lower translation speed due to wasteful processing.

\begin{algorithm}
\begin{algorithmic}
\Procedure{Batching}{encoded sentences $i$}

\State Create batch $b$ from $i$

\While{hypo $h \neq EOS, \forall h \in b $ } 
  \State Decode($b$)
\EndWhile

\EndProcedure

\end{algorithmic}
\caption{Na\"ive mini-batching}
\label{algo:simple-mini-batching}
\end{algorithm}

Amun implements an efficient batching algorithm that takes into account the actual number of hypotheses that need to be decoded at each decoding step, Figure~\ref{algo:Mini-batching}.

\begin{algorithm}
\begin{algorithmic}
\Procedure{Batching}{encoded sentences $i$}

\State Create batch $b$ from $i$

\While{$b \neq \emptyset$} 
  \State Decode($b$)
  
  \ForAll{hypo $h \in b$}
    \If{$h = EOS$  }
      \State Remove $h$ from $b$
    \EndIf
  \EndFor
\EndWhile

\EndProcedure

\end{algorithmic}
\caption{Mini-batching}
\label{algo:Mini-batching}
\end{algorithm}

We will compare the effect of the two implementations in the Section~\ref{sec:Result}.

\subsection{Softmax and K-Best Fusion}

Most NMT models predict a large number of classes in their output layer, corresponding to the number of words or subword units in their target language. For example, ~\citet{sennrich-haddow-birch:2016:P16-12} experimented with target vocabulary sizes of 60,000 and 90,000 sub-word units.

The output layer of most deep learning models consist of the following steps
\begin{enumerate}
   \item \vspace{-2 mm} multiplication of the weight matrix with the input vector $p = w x$
   \item \vspace{-2 mm} addition of a bias term to the resulting scores $p = p + b$
   \item \vspace{-2 mm} applying the activation function, most commonly softmax $ p_i = \exp(p_i) / \sum \exp(p_i) $
   \item \vspace{-2 mm} a search for the best (or k-best) output classes $\argmax_i p_i$
\end{enumerate}

\begin{figure}
\centering
\begin{tabular}{cc}
{\includegraphics[scale=0.35]{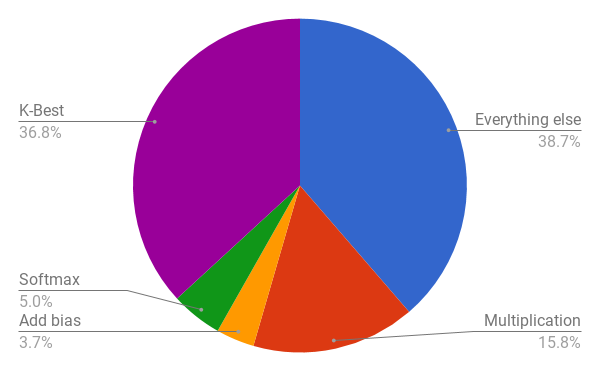}} 
\end{tabular}
\caption{Proportion of time spent during translation}
\label{fig:pie-time}
\end{figure}

Figure~\ref{fig:pie-time} shows the amount of time spent in each step during translation. Clearly, the output layer of NMT models are very computationally expensive, accounting for over 60\% of the translation time.

We focus on the last three steps; their outline is shown in Algorithm~\ref{algo:original-softmax-beamsearch}. For brevity, we show the algorithm for 1-best, a k-best search is a simple extension of this.

\begin{algorithm} 
\begin{algorithmic}

\Procedure{AddBias}{vector $p$, bias vector $b$}
\ForAll{$p_i$ in $p$}
  \State $p_i \gets p_i + b_i$
\EndFor 
\EndProcedure

\State

\Procedure{Softmax}{vector $p$}

\Comment{calculate max for softmax stability}
\State $max \gets - \infty$ 
\ForAll{$p_i$ in $p$}
  \If{$p_i > max$}
    \State $max \gets p_i$
  \EndIf
\EndFor

\Comment{calculate denominator} 
\State $sum \gets 0$ 
\ForAll{$p_i$ in $p$}
  \State $sum \gets sum + \exp(p_i - max)$
\EndFor

\Comment{calculate softmax}
\ForAll{$p_i$ in $p$}
  \State $p_i \gets \frac{\exp(p_i - max)}{sum} $
\EndFor 

\EndProcedure

\State

\Procedure{Find-Best}{vector $p$}

\State $max \gets - \infty$ 
\ForAll{$p_i$ in $p$}
  \If{$p_i > max$}
    \State $max \gets p_i$
    \State $best \gets i$
  \EndIf
\EndFor 

\Return $max$, $best$

\EndProcedure

\end{algorithmic}
\caption{Original softmax and k-best algorithm}
\label{algo:original-softmax-beamsearch}
\end{algorithm}

As can be seen, the vector p is iterated over five times - once to add the bias, three times to calculate the softmax, and once to search for the best classes. We propose fusing the three functions into one kernel, a popular optimization technique~\citep{Guevara2009EnablingTP}, making use of the following observations.

Firstly, softmax and $\exp$ are monotonic functions, therefore, we can move the search for the best class from FIND-BEST to SOFTMAX, at the start of the kernel.

Secondly, we are only interested in the probabilities of the best classes during inference, not of all classes. Since they are now known at the start of the softmax kernel, we compute softmax only for those classes.

\begin{algorithm}
\begin{algorithmic}
\Procedure{Fused-Kernel}{vector $p$, bias vector $b$}

%\Comment{add bias, calculate $max$ \& $argmax$}

\State $max \gets - \infty$ 
\State $sum \gets 0$ 

\ForAll{$p_i$ in $p$}
  \State $p_i' \gets p_i + b_i$  
  \If{$p_i' > max$}
    \State $\Delta \gets max - p_i'$
    \State $sum \gets \Delta \times sum + 1 $
    \State $max \gets p_i'$
    \State $best \gets i$
  \Else
    \State $sum \gets sum + \exp(p_i' - max)$
  \EndIf
\EndFor

\Return $\frac{1}{sum}$, $best$ 

\EndProcedure
\end{algorithmic}

\caption{Fused softmax and k-best}
\label{algo:Fused Kernel}
\end{algorithm}

Thirdly, the calculation of $max$ and $sum$ can be accomplished in one loop by adjusting $sum$ whenever a higher $max$ is found during the looping:
\begin{eqnarray*}
sum & = & e^{x_t - max_b} + \sum_{i=0...t-1}{e^{x_i - max_b}} \\
    & = & e^{x_t - max_b} + \sum_{i=0...t-1}{e^{x_i - max_a + \Delta}}  \\
    & = & e^{x_t - max_b} + e^{\Delta} \times \sum_{i=0...t-1}{e^{x_i - max_a}} 
\end{eqnarray*}

where $max_a$ is the previous maximum value, $max_b$ is the now higher maximum value, i.e., $max_b > max_a$, and $\Delta = max_a - max_b$. The outline of our function is shown in Algorithm~\ref{algo:Fused Kernel}.

In fact, a well known optimization is to skip softmax altogether and calculate the argmax over the input vector, Algorithm~\ref{algo:Argmax only}. This is only possible for beam size 1 and when we are not interested in returning the softmax probabilities.

\begin{algorithm}
\begin{algorithmic}
\Procedure{Fused-kernel-1-best}{vector $p$, bias vector $b$}

\State $max \gets - \infty$ 
\ForAll{$p_i$ in $p$}
  \If{$p_i + b_i > max$}
    \State $max \gets p_i + b_i$
    \State $best \gets i$
  \EndIf
\EndFor

\Return $best$ 

\EndProcedure

\end{algorithmic}
\caption{Find 1-best only}
\label{algo:Argmax only}
\end{algorithm}

Since we are working on GPU optimization, it is essential to make full use of the many GPU cores available. This is accomplished by well-known parallelization methods which multi-thread the algorithms. For example, Algorithm~\ref{algo:Argmax only} is parallelized by sharding the vector $p$ and calculating $best$ and $max$ on each shard in parallel. The ultimate $best$ is found in the following reduction step, Algorithm~\ref{algo:Parallel Argmax only}.

\begin{algorithm}
\begin{algorithmic}
\Procedure{Fused-kernel-1-best}{vector $p$, bias vector $b$}

\Comment{parallelize}
\State Create shards ${p^1...p^n}$ from $p$
\ParFor{$p^j \in {p^1...p^n}$}
  \State $max^j \gets - \infty$ 
  \ForAll{$p_i^j$ in $p^j$}
    \If{$p_i^j + b_i > max$}
      \State $max^j \gets p_i^j + b_i$
      \State $best^j \gets i$
    \EndIf
  \EndFor

\EndParFor

\Comment{reduce}
\State $max \gets - \infty$ 
\ForAll{$max^j \in max^1...max^n$}
  \If{$max^j > max$}
    \State $max \gets max^j$
    \State $best \gets best^j$
  \EndIf
\EndFor

\Return $best$ 

\EndProcedure

\end{algorithmic}
\caption{Parallel find 1-best only}
\label{algo:Parallel Argmax only}
\end{algorithm}

%These changes do not reduce the asymptotic runtime of our algorithm, which remains at $\mathcal{O}(|p|)$, but it does bring the number of iterations over $p$ down from 5 to 1. In practise, this has a significant affect on inference speed.

\subsection{Half-Precision}

Reducing the number of bits needed to store floating point values from 32-bits to 16-bits promises to increase translation speed through faster calculations and reduced bandwidth usage. 16-bit floating point operations are supported by the GPU hardware and software available in the shared task.

In practise, however, efficiently using half-precision value requires a comprehensive redevelopment of the GPU code. We therefore make do with using the GPU's Tensor Core\footnote{https://devblogs.nvidia.com/programming-tensor-cores-cuda-9/} fast matrix multiplication routines which transparently converts 32-bit float point input matrices to 16-bit values and output a 32-bit float point product of the inputs.

\section{Experimental Setup}
\label{sec:Experimental Setup}

Both of our submitted systems use a sequence-to-sequence model similar to that described in ~\citet{DBLP:journals/corr/BahdanauCB14}, containing a bidirectional RNN in the encoder and a two-layer RNN in the decoder. We use byte pair encoding~\citep{sennrich-haddow-birch:2016:P16-12} to adjust the vocabulary size.

We used a variety of GPUs to train the models but all testing was done on an Nvidia V100. Translation quality was measured using BLEU, specifically multi-bleu as found in the Moses toolkit\footnote{https://github.com/moses-smt/mosesdecoder}. The validation and test sets provided by the shared task organisers were used to measure translation quality, but a 50,000 sentence subset of the training data was used to measure translation speed to obtain longer, more accurate measurements.

\subsection{GRU-based system}

%Our first system uses gated recurrent units (GRU) throughout, trained with Marian~\citep{junczys2018marian}, but submitted both systems using the Amun inference engine.

Our first submitted system uses gated recurred units (GRU) throughout. It was trained using Marian~\citep{junczys2018marian}, but Amun was chosen as inference engine.

We experimented with varying the vocabulary size and the RNN state size before settling for a vocabulary size of 30,000 (for both source and target language) and 256 for the state size, Table~\ref{tab:BLEU for newstest2014}.

\begin{table}
\begin{center}
\begin{tabular}{|l|r|r|r|} \hline
		& \multicolumn{3}{|c|}{\emph{State dim}}	\\ \hline	
Vocab size	& 256	& 512	& 1024 \\ \hline
1.000 		& 12.23 &	& 12.77 \\ 
5,000		& 16.79	&  	& 17.16 \\ 
10,000		& 18.00	& 	& 18.19 \\
20,000		& -	&	& 19.52 \\ 
30,000		& 18.51	& 19.17	& 19.64 \\ \hline
\end{tabular}
\end{center}
\caption{Validation set BLEU (newstest2014) for GRU-based model}
\label{tab:BLEU for newstest2014}
\end{table}

After further experimentation, we decided to use sentence length normalization and NVidia's Tensor Core matrix multiplication which increased translation quality as well as translation speed. The beam was kept at 1 throughout for the fastest possible inference.

\subsection{mLSTM-based system}

Our second system uses multiplicative-LSTM~\citep{krause2017} in the encoder and the first layer of a decder, and a GRU in the second layer, trained with an extension of the Nematus~\citep{sennrich-EtAl:2017:EACLDemo} toolkit which supports such models; multiplicative-LSTM's suitability for use in NMT models has been previously demonstrated by~\citet{pinnis2017}.
As with our first submission, Amun is used as inference engine.
We trained 2 systems with differing vocabulary sizes and varied the beam sizes, and chose the configuration that produced the best results for translation quality on the validation set, Table~\ref{tab:mLSTM BLEU - vocab sizes}. 

\begin{table}
\begin{center}
\begin{tabular}{|l|r|r|} \hline
		& \multicolumn{2}{|l|}{\emph{Vocab size}}	\\ \hline	
Beam size	& 40,000	& 50,000 \\ \hline
1 		& 23.45		& 23.32	\\ 
2		& 24.15		& 24.04	\\
3		& 24.48		&  	\\
4		& 24.42		& 	\\
5		& 24.48		& 	\\ \hline
\end{tabular}
\end{center}
\caption{Validation set BLEU for mLSTM-based model}
\label{tab:mLSTM BLEU - vocab sizes}
\end{table}

\section{Result}
\label{sec:Result}

\subsection{Batching}

\begin{figure}
\centering
\begin{tabular}{cc}
{\includegraphics[scale=0.35]{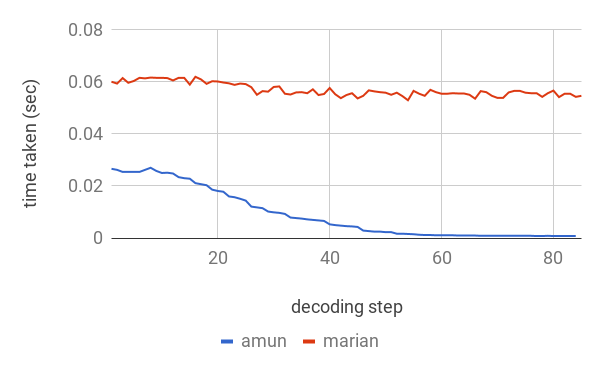}} 
\end{tabular}
\caption{Time taken for each decoding step for a batch of 1280 sentences}
\label{fig:decode-step}
\end{figure}

The efficiency of Amun's batching algorithm can be seen by observing the time taken for each decoding step in a batch of sentences, Figure~\ref{fig:decode-step}. Amun's decoding becomes faster as sentences are completely translated. This contrasts with the Marian inference engine, which uses a na\"ive batching algorithm, where the speed stays relatively constant throughout the decoding of the batch.

\begin{figure}
\centering
\begin{tabular}{cc}
{\includegraphics[scale=0.35]{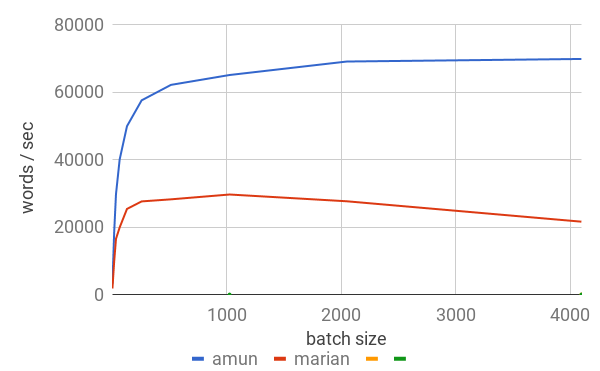}} 
\end{tabular}
\caption{Speed v. batch size}
\label{fig:batch-size}
\end{figure}

Using batching can increase the translation speed by over 20 times in Amun, Figure~\ref{fig:batch-size}. Just as important, it doesn't suffer degradation with large batch sizes, unlike the na\"ive algorithm which slows down when batch sizes over 1000 is used. This scalability issue is likely to become more relevant as newer GPUs with ever increasing core counts are released.
%increase in importance with new GPUs with ever increasing core counts.

\subsection{Softmax and K-Best Fusion}

\begin{figure}
\centering
\begin{tabular}{cc}
{\includegraphics[scale=0.35]{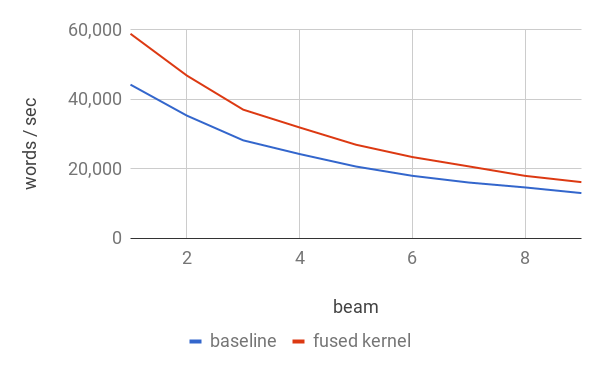}} 
\end{tabular}
\caption{Using fused operation}
\label{fig:fused}
\end{figure}

Fusing the bias and softmax operations in the output layer with the beam search results in a speed improvement by 25\%, Figure~\ref{fig:fused}. Its relative improvement decreases marginally as the beam size increases.

Further insight can be gained by examining the time taken for each step in the output layer and beam search, Table~\ref{tab:fused-breakdown}. The fused operation only has to loop through the large cost matrix once, therefore, for low beam sizes its is comparable in speed to the simple kernel to add the bias. For higher beam sizes, the cost of maintaining the n-best list is begins to impact on speed.

\begin{table}[h]
\begin{center}
\begin{tabular}{|l|r|r|} \hline
		& Baseline	& Fused \\ \hline
\multicolumn{3}{|c|}{\emph{Beam size 1}}	\\ \hline	
Multiplication 	& 5.39 		& 5.38 (+0\%) \\ \hline
Add bias 	& 1.26		&  \\ 
Softmax 	& 1.69		& 2.07 (-86.6\%)\\
K-best extr.	& 12.53		&  \\ \hline
\multicolumn{3}{|c|}{\emph{Beam size 3}}	\\ \hline	
Multiplication 	& 14.18 	& 14.16 (+0\%) \\ \hline
Add bias 	& 3.76		&  \\ 
Softmax 	& 4.75		& 3.43 (-87.1\%)\\
K-best extr.	& 18.23		&  \\ \hline
\multicolumn{3}{|c|}{\emph{Beam size 9}}	\\ \hline	
Multiplication 	& 38.35 	& 38.42 (+0\%) \\ \hline
Add bias 	& 11.64		&  \\ 
Softmax 	& 14.4		& 17.5 (-72.1\%)\\
K-best extr.	& 36.7		&  \\ \hline
\end{tabular}
\end{center}
\caption{Time taken (sec) breakdown}
\label{tab:fused-breakdown}
\end{table}

\subsection{Tensor Cores}

By taking advantage of the GPU's hardware accelerated matrix multiplication, we can gain up to 20\% in speed, Table~\ref{tab:tensor-cores}.

\begin{table}[h]
\begin{center}
\begin{tabular}{|l|r|r|} \hline
Beam size	& Baseline	& Tensor Cores \\ \hline
1	 	& 39.97 	& 34.54 (-13.6\%) \\ 
9 		& 145.8		& 116.8 (-20.0\%) \\ \hline
\end{tabular}
\end{center}
\caption{Time taken (sec) using Tensor Cores}
\label{tab:tensor-cores}
\end{table}

\section{Conclusion and Future Work}

We have presented some of the improvement to Amun which are focused on improving NMT inference.
%to improvement NMT inference.

We are also working to make deep-learning faster using more specialised hardware such as FPGAs. It would be interesting as future work to bring our focused approach to fast deep-learning inference to a more general toolkit. 

%\section*{Acknowledgments}
%This work is sponsored by the Air Force Research Laboratory, prime contract FA8650-11-C-6160.  The views and conclusions contained in this document are those of the authors and should not be interpreted as representative of the official policies, either expressed or implied, of the Air Force Research Laboratory or the U.S. Government.

\bibliographystyle{apalike}
\bibliography{amta2016,mt,more}

\end{document}